\documentclass[11pt]{article}
\usepackage{acl2015}
\usepackage{times}
\usepackage{url}
\usepackage{latexsym}
\usepackage[utf8]{inputenc}
\usepackage{microtype}

\title{Social Media Argumentation Mining:\\The Quest for Deliberateness in Raucousness\thanks{\ \,Extended abstract of a talk presented at the Dagstuhl Seminar 16161: \emph{Natural Language Argumentation: Mining, Processing, and Reasoning over Textual Arguments}, Dagstuhl, Germany, April 2016.}}

\author{Jan Šnajder\\
  University of Zagreb, Faculty of Electrical Engineering and Computing \\
  Text Analysis and Knowledge Engineering Lab\\
  Zagreb, Croatia\\
  {\tt jan.snajder@fer.hr}}

\date{}

\begin{document}

\maketitle

\begin{abstract}
Argumentation mining from social media content has attracted increasing
attention. The task is both challenging and rewarding. The informal nature of
user-generated content makes the task dauntingly difficult. On the other hand,
the insights that could be gained by a large-scale analysis of social media
argumentation make it a very worthwhile task. In this position paper I discuss
the motivation for social media argumentation mining, as well as the tasks and
challenges involved.
\end{abstract}

\section{Introduction}

Argumentation mining aims to detect the argumentative discourse structure in
text and recognize the components of an argument and relations between them.
It is an emerging and exciting field at the confluence of natural language
processing (NLP), logic-based reasoning, and argumentation theory; see
\cite{moens2014argumentation,lippi2015argumentation} for a comprehensive and
recent overview.  

While computational approaches to argumentation already have a long-standing
tradition within the field of artificial intelligence, in particular in
research on logic-based reasoning and multi-agent systems, it is only in recent
years that argumentation mining has begun to attract attention of the NLP
community. From an NLP perspective, argumentation mining is a daunting task,
one that involves many levels of semantic processing (ranging from lexical
semantics to discourse-level processing), and essentially calls for text
understanding and inference mechanisms that significantly surpass the state of
the art.

This is of course not to say that significant advances in the processing of
natural language arguments cannot already be made -- as a matter of fact, the
argumentation mining community has made a significant progress in recent years.
Moreover, the community now has a much better understanding of the set of tasks
involved in argumentation mining, as well as their complexity.

In this short position paper, I focus on argumentation mining in the context of social
media, more specifically opinionated user comments. My aim is
twofold: first, to present possible motivations for argumentation in social
media; secondly, to outline some of the tasks and challenges involved.

\section{Argumentation in Social Media}

Initial work on argumentation mining has focused on well-structured, edited
text, such as legal text \cite{walton2005argumentation} or scientific
publications \cite{jimenez2007argumentation}. Obviously, such genres are
interesting as they exhibit all the characteristics argumentation theory is
concerned with. At the same time, edited text is amenable to NLP as it can be
processed with existing, mostly non-robust tools.

Recently, however, the focus has also shifted to argumentation mining from
social media texts,  such as online debates
\cite{cabrio2012combining,habernal2014argumentation,boltuzic2014back},
discussions on regulations \cite{park2014identifying}, and product reviews
\cite{ghosh2014analyzing}.

Online debates are particularly well-suited for argumentation mining, because
of the controlled setting offered by online debate platforms, and because most
users will use these platforms with an \emph{intention} to engage in
argumentative discussions. The same cannot be said of less controlled
communication environments, such as comment boards on news portals, product
review sites, or microblogs, where the communicative intention is not to engage
in an argumentative discussion, but rather to express a blunt opinion on the
subject matter, or even simply to satisfy the need for self-presentation.

In what follows, I use the term \emph{opinionated comment} to refer to such
user-generated content, one which is not necessarily generated within a debate.
As a prototypical example is a comment on a news article. As an example,
consider the following opinionated comment related to the \emph{Trump rally}
event:\footnote{Yahoo News, \url{http://tinyurl.com/zkez7ze}}

\begin{quote}
\emph{%
The President we have now divided our country and put his ego first instead of the people. Trump hasn't divided the country that's why he has so many people behind him. We want someone who is not afraid of the politics in Washington and change our policies with dealing with other countries.}
\end{quote}

Clearly, the author of this comment does express some arguments to back up his or
her opinion. However, the opinion is triggered by an event, and there is no
predefined debate topic. Moreover, there will likely be no follow-up discussion
in which the author would need to justify or elaborate on his or her arguments.
Thus, it seems that such opinionated comments mostly emerge ad hoc and are
monological in nature. 

It is legitimate to ask whether there is any merit in analyzing this kind of
opinionated text, apart from the fact that it is abundant in social media. I
argue that -- to an extent in which we are interested in analyzing the opinions
of other people (users of products, voters, etc.) -- we should also be
interested in analyzing the \emph{reasons} underpinning those opinions, for
otherwise we cannot fully apprehend them. If you are, say, running a political
campaign, you would want to know what people think of you and \emph{why}. You
would probably also want to do this analysis across all the events that are
even marginally related to your campaign, and you would also want to do it on a
large scale to get the ``totality of the experience.''

\section{Challenges}

Besides the challenges mentioned in the introduction, there is a number of
additional challenges involved in argumentation mining from user-generated
text:

\begin{enumerate}
\item \textbf{Noisy text.} \newcite{baldwin2013noisy} demonstrate that social media sources are more noisy than edited texts, although they can also be cleaned using NLP techniques;

\item \textbf{Vague claims.} It is  probably safe to say that the majority of
online users do not really see a need to present a well-formed argument for
their position.   As a consequence, claims made by the
users will often be unclear, ambiguous, vague, or simply poorly worded.  This is
even the case for more discussion-tuned environments, such as online debate
platforms.

\item \textbf{Vague argument structure.} Again, because the users rarely feel
the need to argue for their position, most user-generated opinionated text will
not constitute a properly structured argument. This is especially true for
short texts, such as microbloging posts. Even when there are some traces of an argumentative
structure, it will likely be incomplete and lack important premises.

\end{enumerate}

\section{Tasks}

A number of argumentation tasks have been proposed in the literature. The two main ones are:

\begin{itemize}
\item \textbf{Component identification} -- the task of detecting the premises and conclusion of an argument, as found in a text of discourse;
\item \textbf{Relation prediction} -- identifying the relations between components.
\end{itemize}

In a recent study on user-generated social media texts,
\newcite{habernal2016argumentation} showed that (a slightly modified) Toulmin
model of argumentation may be suitable for short documents, such as article
comments or forum posts. They annotated the claim, premise, backing, rebuttal,
and refutation components, thereby achieving a moderate inter-annotator
agreement. They use sequence labeling to tackle the component identification
task, reaching a token-level F1-score of 0.25.

Component identification and relation prediction are without a doubt relevant
argumentation mining tasks. However, coming back to the political campaign
example, it is not immediately obvious how these tasks can aid in analyzing the
reasons underpinning the opinions, especially when dealing with large volumes
of data. To analyze arguments on a large scale, it seems that we at least need
to:

\begin{itemize}
\item \textbf{Identify the main arguments} -- identify the main (central, most prominent, most often used) arguments that the users use when discussing a certain topic. An \emph{argument} here is meant to mean a claim and (the possibly convergent) set of premises supporting it;
\item \textbf{Classify opinionated posts} -- given an opinionated post, identify its main arguments.
\end{itemize}

Consider again the \emph{Trump rally} example from above. The post may be
classified as belonging to the main argument \emph{``Donald Trump would make a
good president''}. The main claim is \emph{``Donald Trump will change the
foreign policy for the better''}, while the supporting premises may be
\emph{``Existing foreign policy is bad''}, \emph{``Trump is not afraid to take
on the Establishment''}, etc.

Given a large-enough amount of user-generated opinionated data, there seem to
be at least two ways in which main arguments could be identified. First, the
arguments could be extracted manually. This is essentially what we have done in
\cite{boltuzic2014back}, where we used the main claims distilled from an online
debating platform. Similarly, \newcite{hasan2014why} asked annotators to group
the user comments and identify the main claims. The second option is to resort
to unsupervised machine learning and try to induce the main arguments (or at
least the main claims) automatically, in a bottom-up fashion. A middle-ground
solution, proposed by \newcite{sobhani2015argumentation}, is to use
unsupervised machine learning to induce the argument clusters, and then map
those clusters manually to main arguments.

From a machine learning perspective, the two above-mentioned tasks may be framed as follows:

\begin{itemize}
\item \textbf{Argument clustering} -- grouping of similar arguments, so that the main arguments/claims can be identified;
\item \textbf{Argument classification} -- given an opinionated comment, classify it into one or many classes, each corresponding to one main argument (obtained either manually or using argument clustering).
\end{itemize}

In \cite{boltuzic2015identifying}, we tackled the former task and 
investigated the suitability of semantic textual similarity (STS) \cite{agirre2012semeval} for clustering the
main claims. Our conclusion was that fully automatic argument clustering is
hardly feasible, however we hypothesized that it might prove valuable in a
computer-aided or semi-supervised argumentation mining setup.

We tackled the task of argument classification in \cite{boltuzic2014back},
under the name ``argument recognition'', while \newcite{hasan2014why} tackled
the same task in the context of stance detection, under the name ``reason
classification''.  The main difference is that \newcite{hasan2014why} frame the
problem as a (joint learning) supervised text classification task with lexical
features, which makes their model topic-specific. In other words, the model
learns to classify the user comments into classes that correspond to main
claims, without explicitly comparing the user comments against the main claims.
In contrast, in \cite{boltuzic2014back} we modeled the similarity between the
user comments and the main claims using STS and textual
entailment predictions and fed these to a supervised model. At least in
principle, this should make the model topic-independent. The model outperformed
the baseline, although not by a large margin.

\section{Argument Similarity}

What the two tasks above have in common (with the exception of argument classification using
lexical features, which is topic-specific and hence arguably the least
practical approach) is the requirement to compute the similarity between two
arguments. The \emph{argument similarity} has been introduced in
\cite{boltuzic2015identifying}, as well as in
\cite{swanson2015argument,misra2015using}, under the name \emph{argument facet
similarity}. Intuitively, a pair of arguments should receive the highest score
if they mean the same, and the lowest score if they are completely
different and on a different topic. Ideally, argument similarity would account for 
both the similarity of argument components as well as the similarity of
argument structures (how components relate to each other). 

\newcite{misra2015using} consider the similarity between main claims expressed
in user-generated arguments. They develop a regression model using a number of
comparison features, including STS. Their model, trained on human-annotated
pairs of claims, reaches a correlation score of 0.54, outperforming a sensible
baseline. In contrast, in \cite{boltuzic2015identifying} we model argument
similarity on the dataset from \cite{hasan2014why}, in an unsupervised fashion,
using word embedding representations.

\section{Nano-level Argument Processing}

Work cited above seems to indicate that the existing approaches to measuring
STS provide only limited means to measure argument similarity. Consider the
following examples on the \emph{Marijuana legalization} topic from the
\cite{hasan2014why} dataset:

\begin{itemize}
\setlength\itemsep{0.1em}
\item Comment 1: \emph{Legalizing marijuana could potentially lower the number of users.}
\item Comment 2: \emph{Now it is not taxed, and those who sell it are usually criminals of some sort (though many are harmless).}
\item Main claim: \emph{Legalized marijuana can be controlled and regulated by the government.}
\end{itemize}

In this case, both comments have been classified by human annotators as
essentially expressing the main claim (one of the main claims identified by
analyzing the complete dataset). However, at first glance it is not obvious how
these two comments are \emph{similar} to each other or to the main claim. It is
also very unlikely that they would be predicted as similar by an STS system, given
the large semantic gap between them. However, assuming that the main claim is
indeed the best fit, most people would probably be able to come up with sets
of \emph{implicit premises} holding between each of the two comments and the main claim. For instance, the following premises link Comment 2 to the main claim:

\begin{itemize}
\setlength\itemsep{0.1em}
\item \emph{If a thing is not taxed, criminals can sell it.}
\item \emph{Criminals should be stopped from selling things.}
\item \emph{Things that are taxed are controlled and regulated by the government.}
\end{itemize}

Current approaches to argument similarity are not generative in nature and
cannot generate a chain of implicit premises.  The task seems to be related to
what \newcite{lippi2015argumentation} refer to as the \emph{completion task}: the
task of inferring implicit argument components. Alternatively, if we take micro-level
argumentation to focus on the components of a single argument, then the
functionality to infer the similarity between two arguments can perhaps be
dubbed \emph{nano-level argumentation}.  While this task has apparently not yet
been addressed in the literature, it seems to be a necessary ingredient of an
argumentation mining system capable of analyzing user-generated arguments on a
large scale.  

\section{Conclusion}

Social media argumentation mining allows us to understand the reasons
underpinning user opinions. However, mining opinionated comments poses a number
of challenges related to the informal and non-structured nature of
user-generated text. Analyzing such text on a large scale calls for the ability
to compute the similarity of arguments, either to identify the main claims or
to classify the arguments by their main claims. A principled solution will
probably have to operate at the \emph{nano-level} of argumentation, i.e., infer
(rather than merely measure) the similarity between two claims.

\bibliographystyle{acl}
\bibliography{arg}

\end{document}